%% file: example_paper.tex
\documentclass{article}

\usepackage{microtype}
\usepackage{graphicx}
\usepackage{booktabs} % for professional tables

\usepackage{xcolor}         % colors

\usepackage{subfig}
\usepackage{amsmath}
\usepackage{xspace}
\usepackage{multicol}
\usepackage{multirow}
\usepackage{tabulary}
\newcommand{\paper}{\textsc{planer}\xspace}

\usepackage{amsmath}
\usepackage{amssymb}
\usepackage{mathtools}
\usepackage{amsthm}
\usepackage{float}
\usepackage{comment}
\usepackage{hyperref}

\usepackage[utf8]{inputenc}
 
\usepackage[accepted]{mlsys2022}

\mlsystitlerunning{Efficient Sparsely Activated Transformers}

\begin{document}

\twocolumn[
\mlsystitle{Efficient Sparsely Activated Transformers}

% unless you've provided the [accepted] option to the mlsys2022


\mlsyssetsymbol{equal}{*}

\begin{mlsysauthorlist}
\mlsysauthor{Salar Latifi}{umich}
\mlsysauthor{Saurav Muralidharan}{nvidia}
\mlsysauthor{Michael Garland}{nvidia}
\end{mlsysauthorlist}

\mlsysaffiliation{umich}{Department of Computer Science and Engineering, University of Michigan, Ann Arbor, USA}
\mlsysaffiliation{nvidia}{NVIDIA Corporation, Santa Clara, USA}

\mlsyscorrespondingauthor{Salar Latifi}{salar@umich.edu}
\mlsyscorrespondingauthor{Saurav Muralidharan}{sauravm@nvidia.com}

%\mlsyskeywords{Machine Learning, MLSys}

\vskip 0.3in

\begin{abstract}
\input{tex/abstract}
\end{abstract}
]



\printAffiliationsAndNotice{} % otherwise use the standard text.
\thispagestyle{plain}
\pagestyle{plain}

\input{tex/introduction}

\input{tex/background_motivation}
\input{tex/methodology}
\input{tex/results}

\input{tex/related}
\input{tex/conclusion}

\section*{Acknowledgements}
%This material is based upon work supported by DARPA under Contract No. HR0011-18-3-0007. Any opinions, findings and conclusions or recommendations expressed in this material are those of the author(s) and do not necessarily reflect the views of the U.S. Government. Distribution Statement "A" (Approved for Public Release, Distribution Unlimited). 
This research was developed with funding from the Defense Advanced Research Projects Agency (DARPA). The views, opinions and/or findings expressed are those of the author and should not be interpreted as representing the official views or policies of the Department of Defense or the U.S. Government.
Distribution Statement ``A'' (Approved for Public Release, Distribution Unlimited). 
\bibliography{example_paper}
\bibliographystyle{mlsys2022}

\appendix
\onecolumn

\input{tex/appendix}

\end{document}

%% file: tex/abstract.tex
Transformer-based neural networks have achieved state-of-the-art task performance 
in a number of machine learning domains including natural language processing and computer vision.
To further improve their accuracy, recent work has explored the integration of dynamic behavior into 
these networks in the form of mixture-of-expert (MoE) layers. In this paper, we explore the introduction
of MoE layers to optimize a different metric: inference latency. We introduce a novel system 
named \paper that takes an existing Transformer-based network and a user-defined latency target and produces
an optimized, sparsely-activated version of the original network that tries to meet the latency target while 
maintaining baseline accuracy. We evaluate \paper on two real-world language modeling tasks using the Transformer-XL network and achieve inference latency reductions of over 2x at iso-accuracy.

%% file: tex/introduction.tex
\section{Introduction}

\begin{figure}[tb]
  \centering
  \includegraphics[width=1\linewidth]{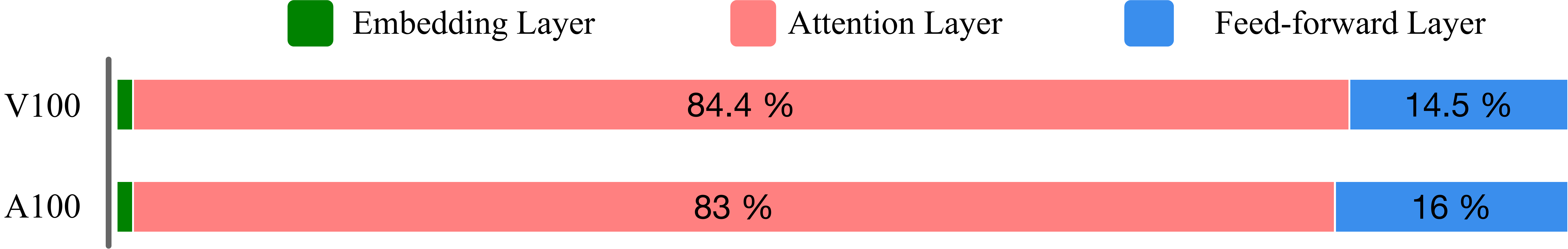}
  \caption{Profiling results for different Transformer-XL layers on NVIDIA V100 and A100 GPUs}
  \label{fig:attention-profile}
\end{figure}

Attention-based deep neural networks (DNNs) such as Transformer~\citep{vaswani2017attention} and BERT~\citep{devlin2018bert} have been shown to exhibit state-of-the-art performance across a variety of machine learning domains, including natural language processing~\citep{wolf2020transformers} and computer vision~\citep{dosovitskiy2020image}. Due to their size and complexity, they are expensive to train and deploy, especially on resource-constrained hardware. In particular, attention layers, which form the building blocks of such networks, account for the majority of network runtime.
Figure~\ref{fig:attention-profile} illustrates this using the Transformer-XL network~\citep{dai2019transformer}; here, we show the proportion of inference latency that each layer type is responsible for on two different GPUs: the NVIDIA V100 and NVIDIA A100. We notice that on both GPUs, attention layers (shown in red) account for over 80\% of total inference latency, with the rest coming from feed-forward (blue) and embedding layers (green).
Due to their outsize influence on total inference latency, recent work has explored various approaches for runtime performance optimization that specifically target attention layers; this includes work such as PAR Transformer~\citep{mandava2020pay}, where attention layers are re-distributed within the network to optimize performance, and various papers on pruning either attention heads and/or entire attention layers~\citep{wang2020hat}.
  
%* However, they are expensive to train and inference can be slow.
%* Specifically, attention layers are expensive and are one of the main reasons
%  why Transformer-based networks are slow.
%  * Chart showing breakup of runtimes by layer type
%  * 2 stacked bars, one for V100 and A100
%* The distribution of FFN-Attention layers may not be optimal and can be optimized
%  further [Par, Sandwich]

A separate body of work has explored the addition of sparsely activated layers to Transformer models to improve task performance~\citep{shazeer2017outrageously}. In particular, mixture-of-expert (MoE) Transformer variants such as Switch Transformer~\citep{fedus2021switch} have demonstrated state-of-the-art task performance while simultaneously improving training and inference costs. While most work in this direction has focused on improving task accuracy, in this paper we attempt to answer the following question: \emph{can the addition of sparsely activated layers help \textbf{preserve} accuracy in the face of latency-optimizing network transformations such as skipping/pruning attention layers? And if so, to what extent?}

\begin{comment}
* Recent work has explored the addition of dynamic layers to Transformer networks [Switch  
  Transformer, others]
* Why? Input space partitioning. [Get some material from existing MoE papers]
* Here, an FFN is replaced by an MoE FFN. This has largely been done to improve accuracy.
* In this paper, we make the observation that adding MoE layers can also help **preserve**
  accuracy in the face of other transformations targeting runtime speedups.
\end{comment}

\begin{figure}[tb]
  \centering
  \includegraphics[width=1\linewidth]{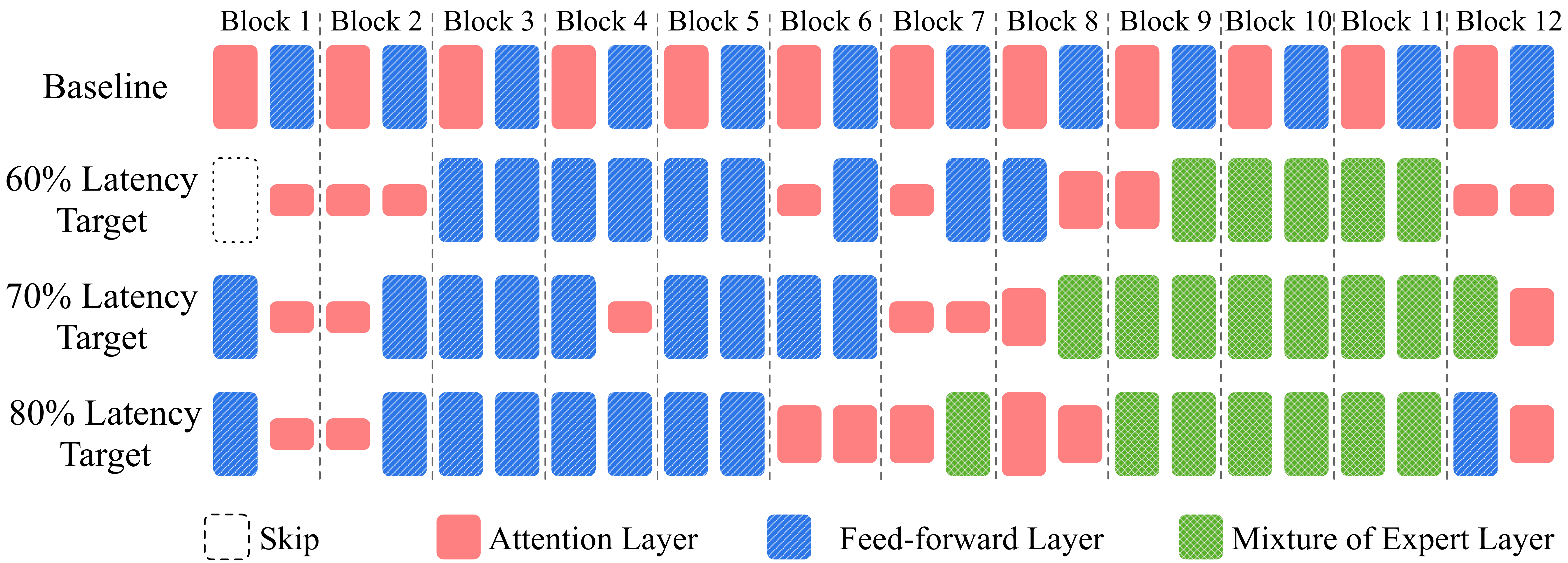}
  \caption{Exploration results for Transformer-XL Base model on enwik8 dataset for different latency targets.}
  \label{introduction_results}
\end{figure}

To help answer this question, we present \paper, a novel system for designing latency-aware sparsely activated Transformer networks.
Given a Transformer-based model as input, along with an inference latency target expressed as a percentage of the baseline model's latency, \paper 
produces a sparsely-activated Transformer model that fulfills the latency objective while preserving baseline accuracy. 
\paper employs an efficient two-phase gradient descent-based neural architecture search (NAS) strategy with a dynamic loss formulation to achieve this.
During the search process, \paper efficiently explores the large number of alternative architectures arising from different combinations of feed-forward, attention (with varying number of heads), and mixture-of-expert layers; as a concrete example, \paper considers over 68 billion unique architectures for the Transformer-XL model in our evaluation.
The optimized architecture obtained from NAS is then fine-tuned using a load-balancing loss term to produce the final network.
Figure~\ref{introduction_results} demonstrates how \paper infers different architectures depending on the user-provided inference latency targets. Here, each of the inferred architectures matches baseline accuracy, but has different inference latencies. Depending on the latency target, we notice that \paper progressively reduces the number of attention layers and their widths, while using additional MoE and/or feed forward layers to compensate for potential accuracy drops.

We evaluate \paper on two different Transformer-based networks drawn from language modeling, and demonstrate an inference latency reduction of at least $2\times$ for each network while maintaining baseline accuracy.
We also compare \paper with prior work such as PAR Transformer~\citep{mandava2020pay} and Sandwich Transformer~\citep{press2019improving}, and with parameter-matched non-MoE implementations of the final optimized networks.

%% file: tex/background_motivation.tex
\section{Background and Motivation}

Mixture-of-expert (MoE) networks~\citep{masoudnia2014mixture} dynamically partition the input domain so that each sub-network or ``expert'' specializes in one or more input partitions, yielding a sparsely activated network.
Recent work has explored the application of MoE layers to efficiently increase the model capacity of Transformer-based architectures~\citep{shazeer2017outrageously, lepikhin2020gshard, fedus2021switch, he2021fastmoe}. These sparsely-activated architectures are shown to achieve similar accuracy gains without the proportional increase in computation compared to traditional scaling of network parameters~\cite{raffel2019exploring}.
In this work, we focus on applying MoE layers to improve inference latency while maintaining baseline accuracy.

\begin{figure}[tb]
\subfloat[MoE Layer]{\includegraphics[height=0.4\linewidth]{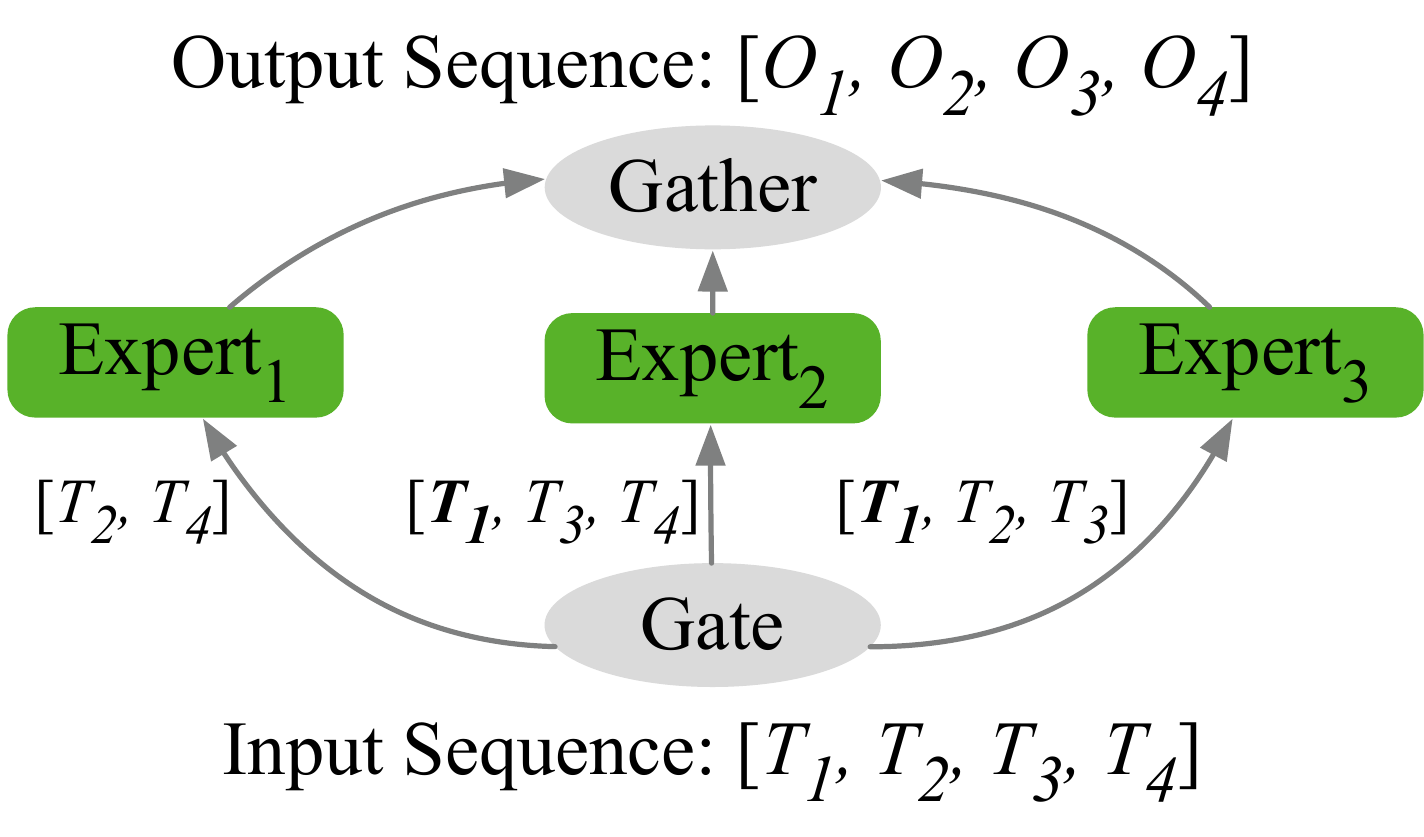}\label{MoE_function}}
\hspace{5mm}
\subfloat[Gate]{\includegraphics[height=0.4\linewidth]{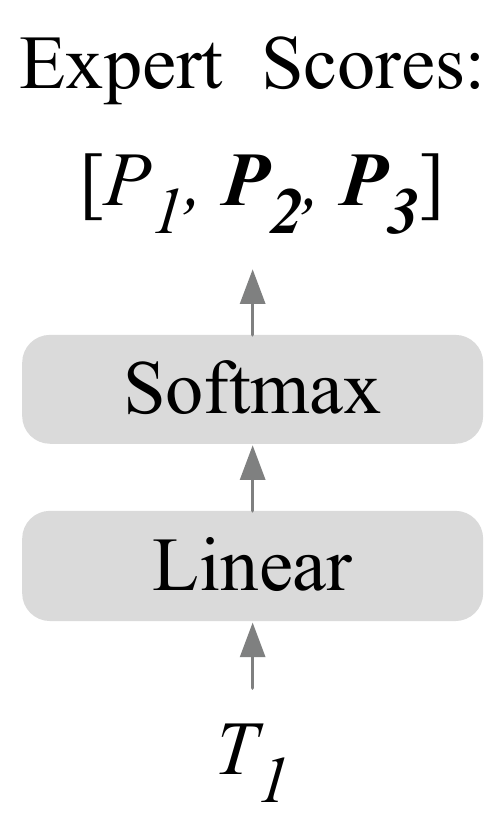}\label{Gate_function}}
\caption{General overview of MoE layers and gate function.}
\end{figure}

Figure~\ref{MoE_function} depicts a general implementation of an MoE layer with three experts. The sequence of input tokens are distributed among the experts for processing, where each token is processed by one or more experts. The number of experts per token is denoted as $Top_K$ in this work. In Figure~\ref{MoE_function}, $Top_K$ is two. A single-layer linear classifier called a \textit{Gate} (Figure~\ref{Gate_function}) decides which expert(s) to use to process a specific token. The Gate generates a probability distribution across the experts per token, which will then be used to select the $Top_K$ experts.

\begin{figure}[tb]
  \centering
  \includegraphics[width=1\linewidth]{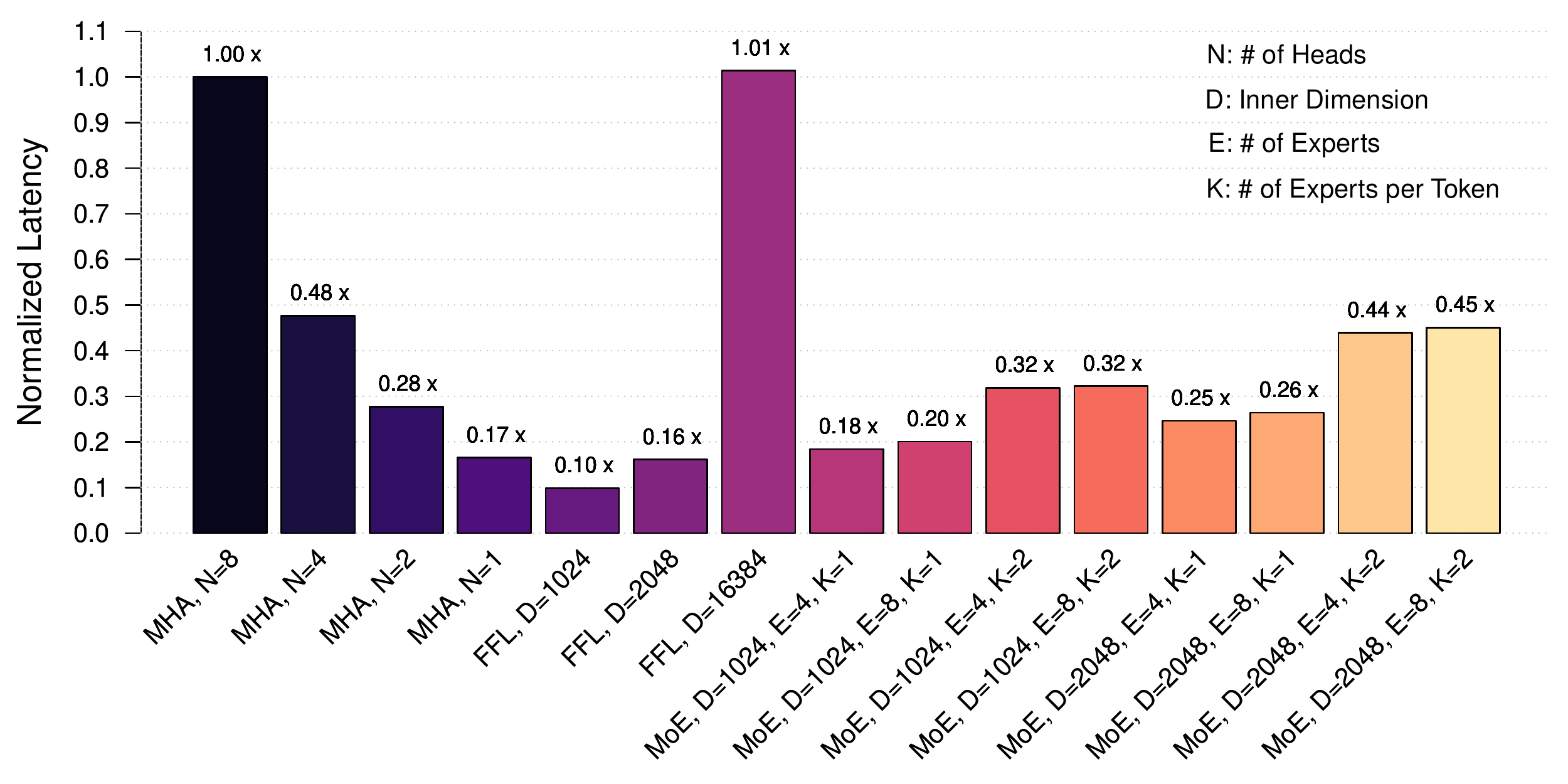}
  \caption{Latency comparison of attention, FFL, and MoE layers normalized w.r.t. attention with 8 heads, profiled on NVIDIA A100 GPU with batch size of 64, sequence length of 192, and half-precision.}
  \label{block_performance}
\end{figure}

\textbf{Layer-wise Performance Analysis:}
To better understand the performance behavior of Transformer-based networks, we present layer-wise profiled latencies for the Transformer-XL Base network in Figure~\ref{block_performance}.
Here, each bar represents the latency of a network block normalized to the latency of default multi-head attention with 8 heads.
Profiling is performed with a model dimension of $512$, $target\_len$ of 64, and batch size of 64 on an NVIDIA A100 GPU. We observe three key points from the figure: (1) the significant cost of the default attention configuration, amounting to a $6.2\times$ higher runtime compared to the default feed-forward layer (FFL) with an inner dimension of 2048, (2) the approximately linear scaling of the attention cost with respect to the number of heads (pruning attention heads and/or blocks could thus play a significant role in improving network performance), and (3) the compute efficiency of the MoE blocks compared to both attention and iso-parametric FFL blocks (iso-parametric FFL blocks are obtained by scaling up an FFL block to match the number of parameters in a corresponding MoE block), signifying the promise of using MoE blocks as a cost-effective solution to compensate for the potential accuracy loss caused by aggressive attention pruning.

%% file: tex/methodology.tex
\section{Searching for Efficient Transformers}
\label{sec:search}

%In this section, we dive into details of the proposed methodology to design a latency-aware transformer architecture. First, we describe the deployed Mixture-of-Expert (MoE) implementation, which plays a vital role in the proposal of efficient architectures. Next, the 2-step search methodology is presented to find the optimal distribution and configuration of the attention layers throughout the network architecture.
In this section, we provide a thorough description of \paper's two-phase NAS methodology for finding optimal latency-aware Transformers.

%To find the optimal distribution and configuration of the attention layers throughout the network architecture, we apply a two-phase search methodology. The first phase is targeted to explore the large design space composed of different configurations of multi-head attentions, FFNs, and MoE layers. The inputs to the first phase are composed of the design space, the backbone of the baseline network architecture, and also a target latency ratio with respect to the baseline latency. 

%After the search phase is finished, the optimized network architecture is instantiated for retraining. This step is necessary for evaluation of the final accuracy level of the selected network, since the building layers have shared the training steps and data during phase 1 with other search options. Therefore, the weights of the selected network at the end of phase 1, are not suitable for the optimal accuracy. There is also an option for fine-tuning the weights from phase 1 in phase 2, but this would require exploring the necessary training hyper-parameters.

\subsection{Phase 1: Search Space Exploration}
Transformer-based models are composed of multiple blocks, where each block consists of multi-head attention (MHA) and feed-forward layers (FFLs)~\citep{vaswani2017attention}.
MoEs could thus be applied to either MHA or FFLs, or both. In this work, we only explore MoE FFLs in the design space; this is primarily due to the runtime overhead introduced by dynamic behavior, which we found to be prohibitively high for the already expensive attention layers.
\paper's first phase explores the large design space composed of different configurations of MHAs, FFLs, and MoE layers. The inputs to the first phase are the design space, the backbone of the baseline network architecture, and a target latency, expressed as a ratio w.r.t. the baseline latency.

For real-world networks, the design space of alternative architectures often gets prohibitively large; for instance, the Transformer-XL Base network on the enwik8 dataset yields a search space size of over 68 billion architectures. To keep the search tractable, we deploy a differentiable NAS strategy, which has been shown to be significantly more efficient than reinforcement-learning-based approaches~\citep{zoph2016neural}. We follow a NAS algorithm similar to the one proposed by~\citet{wu2019fbnet}.

\begin{figure}[t]
  \centering
  \includegraphics[width=1\linewidth]{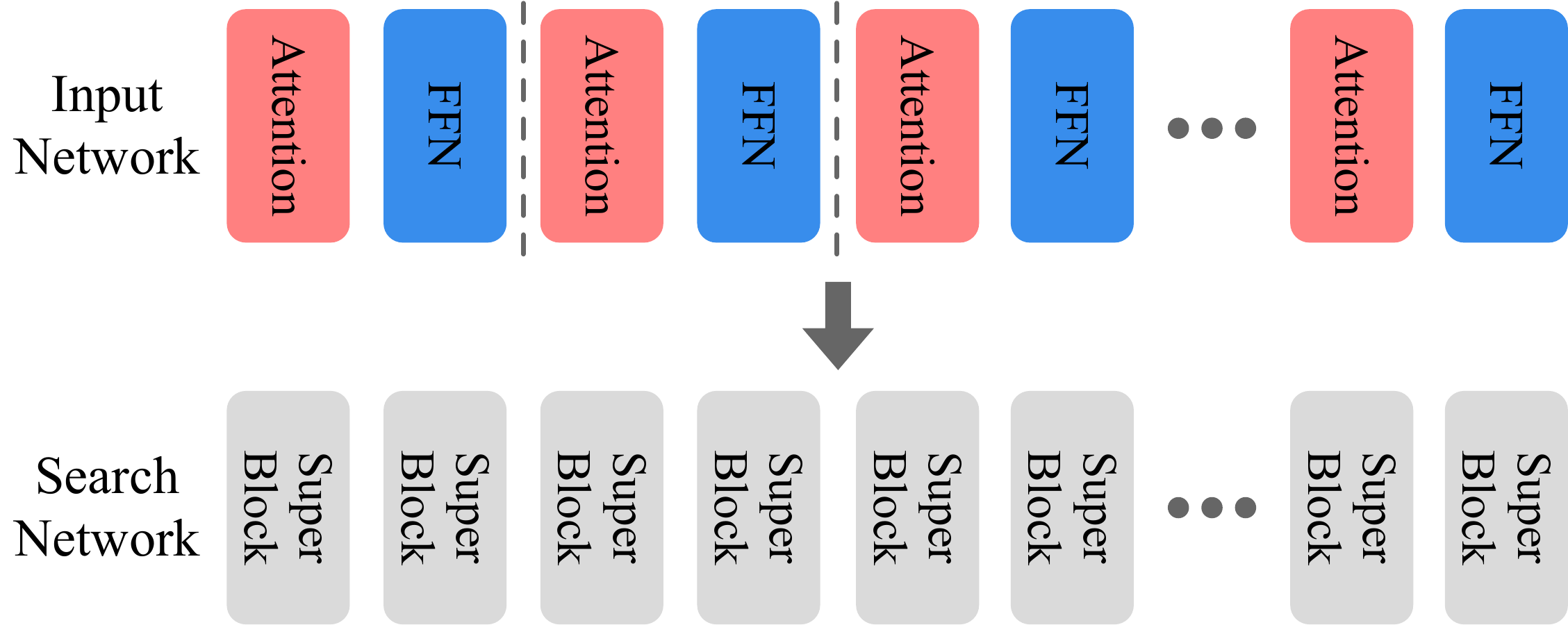}
  \caption{Composing the search network from the input network backbone}
  \label{search_network}
\end{figure}

\begin{figure}[t]
  \centering
  \includegraphics[width=1\linewidth]{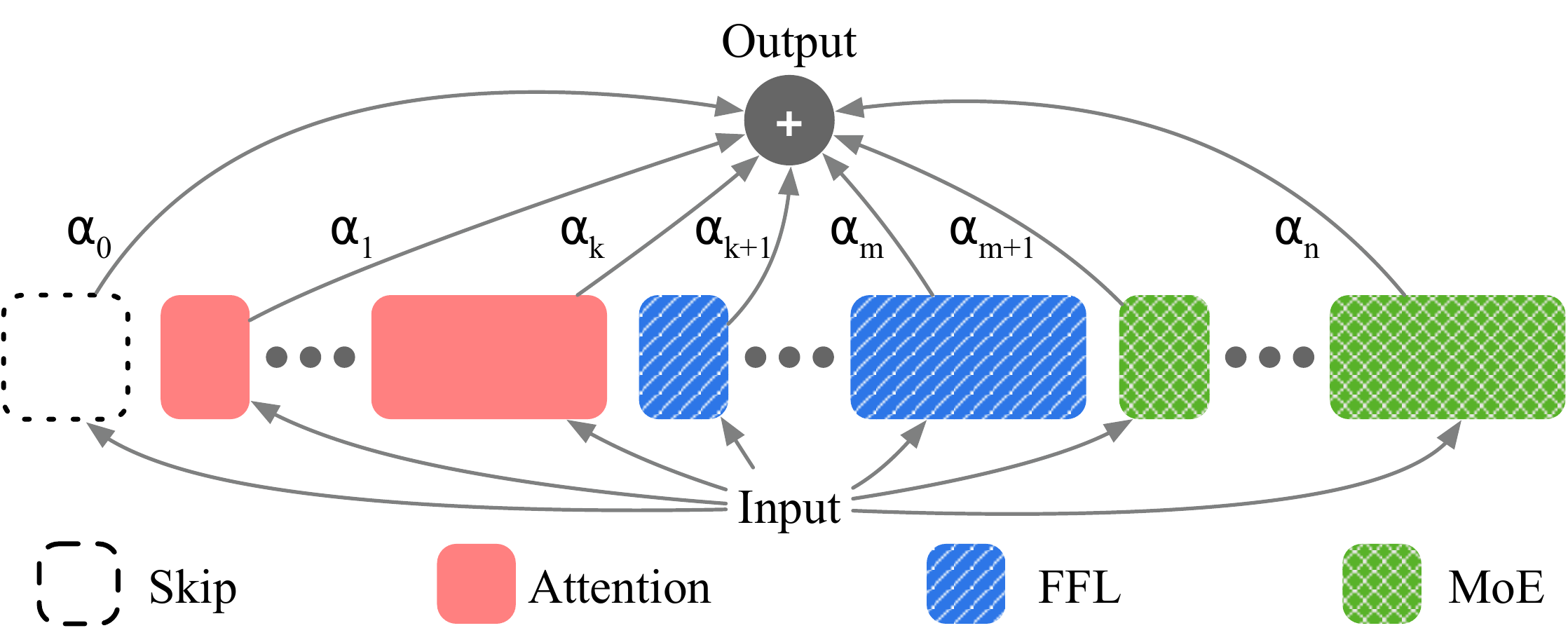}
  \caption{Formulating super blocks from the search space.}
  \label{super_block}
\end{figure}

Phase 1 first composes a \emph{search architecture} using the baseline network's backbone as depicted in Figure~\ref{search_network}. The backbone includes details on the number of blocks (MHA or FFLs) and their configuration (number of heads or hidden dimension). Using the input backbone, each of the MHA or FFL blocks in the baseline network are replaced with \textit{Super Blocks (SB)}, which includes all the search options in the design space. The goal is to find the best option for each block so that overall accuracy is maximized and the latency target is achieved.
Figure~\ref{super_block} depicts the formulation of super blocks. Each of the search options ${Block}_i$ is accompanied by corresponding architectural weights $\alpha_i$, which are trained using gradient descent to represent the benefit factor of the search option~\cite{wu2019fbnet}. To make the optimization graph differentiable with respect to the architecture weights, the output of the super block is formulated as:
\begin{multline}
    Output = \sum_{i=0}^n P_i \times {Block}_i(Input) \\
    s.t. \quad P_i = GumbelSoftmax(\alpha_i, [\alpha_0, ..., \alpha_n]) \quad \
\end{multline}
Where the $GumbelSoftmax$ generates probability values by sampling the Gumbel distribution based on $\alpha$ weights. 

This formulation yields two sets of parameters to be trained in Phase 1. The first group contains the actual network weights (${Block}_i$), and the second group the architectural weights ($\alpha_i$). Training of each parameter group is done sequentially in each epoch, using separate optimizers. Thus each epoch of training in phase 1 consists of optimizing the network weights using $100\%$ of the training samples, and then training the architecture weights using $20\%$ of the randomly sampled training data. We use soft sampling for $GumbelSoftmax$ during architecture optimization, and hard-sampling while training the network weights to reduce the overheads associated with the super blocks. To ensure that neither of the network weight sets are starved due to the hard-sampling of $GumbelSoftmax$, the architecture optimization is initially disabled for $10\%$ of the epochs, and an annealing temperature scheduling is used for later epochs. These settings allow the blocks to be randomly sampled for the appropriate number of search epochs. 

\subsection{User-defined Latency Optimization}
To incorporate latency optimization in the search phase, we formulate an auxiliary loss based on the latencies of the search and baseline network, as well as the target latency. We use an estimation for the end-to-end latency of the search network as well as baseline in phase 1, using lookup tables filled with individual block latencies similar to prior work~\cite{wu2019fbnet}. Equation (2) presents the formulation for the estimated latency which is composed of accumulating the latencies of each super block ($Lat\_SB$).
\begin{equation}
    Lat = \sum_{b=0}^B {Lat\_SB}_b,
    \quad s.t. \quad {Lat\_SB}_b = \sum_{i=0}^n P_{bi} \times {Lat}_i
\end{equation}
Here, ${Lat}_i$ represents the profiled latency of ${Block_i}$ in isolation, and $P_{bi}$ values correspond to the probability values for super block of $b$ as sampled in Equation (1) with respect to the architecture weights. 

The latency loss $Lat_{Loss}$ is implemented as the ratio of the estimated latency of the search network ($Lat$) over the normalized baseline latency with respect to the target latency. 
\begin{multline}
    Loss = {CE}_{Loss} + \beta \times Lat_{Loss} \\
    s.t. \quad Lat_{Loss} = Lat \:\: / \:\:({Lat}_{Baseline} \times {Target}_{Lat}) \\
    s.t. \quad \beta = 1 \quad if \:\: (Lat_{Loss} > 1) \quad else \:\: 0 \qquad \quad \ \:
\end{multline}
During the training of the architecture weights, the latency loss will be automatically activated depending on whether the estimated latency of the search network is meeting the target latency requirement.
For example, if the target latency is set to $50\%$ of the baseline, the latency loss will only get included if the estimated latency is higher than $0.5 \times Lat_{Baseline}$. Otherwise, the scalar factor of $\beta$ would be 0 in Equation 3, leading the optimizer to adjust the architecture weights solely in the direction of minimizing the $CE_{Loss}$. 
This novel dynamic functionality helps the search progress towards the user latency target \emph{without the need for additional hyper-parameter tuning}.

\subsection{Phase 2: Architecture Sampling and Retraining}
The optimized architecture obtained from Phase 1 is now instantiated for retraining. Since the weights of this final architecture were shared with other search points during Phase 1, a retraining step is necessary to avoid under-fitting and to obtain optimal accuracy.
We construct the optimized architecture by selecting the blocks with the highest architecture weight values in each super block; from our empirical evaluation, this sampling strategy best balances additional training overheads with accuracy compared to other approaches such as the one described in ~\citet{liu2018progressive}.
We retrain the sampled architecture from scratch using the same settings as the baseline.

\subsection{Balancing Load Across Experts in MoE Layers}
Since MoE blocks may be part of the final architecture, we incorporate an auxiliary loss during Phase 2 to enforce a balanced load across the experts.
We follow the same implementation of the auxiliary loss for load balancing ($Balance_{Loss}$) as Switch Transformer~\cite{fedus2021switch}. Consider an MoE layer with $E$ experts: 
\begin{multline}
    Loss = {CE}_{Loss} +  Balance_{Loss} \\
    s.t. \quad  Balance_{Loss} = E \times \sum_{e=0}^E {F}_e \times {G}_e \qquad \qquad 
\end{multline}
Here, $F_e$ represents the fraction of the tokens processed by expert $e$, and $G_e$ measures the average gate score received by expert $e$ across the input tokens.

The $Balance_{Loss}$ provides an approximation for the load balancing score across experts. If the tokens are distributed uniformly across the experts by the gate function, we can expect each expert to process $\frac{1}{E}$ of the input tokens, while receiving an average score of $\frac{1}{E}$ from the gate. This would result in $Balance_{Loss}$ having an ideal value of $1$ in a fully-uniform distribution of tokens across the experts. If there is more than one MoE layer in the architecture, the $Balance_{Loss}$ is the average of the individual loss values across the MoE layers.

\begin{figure}[tb]
\subfloat[Comparison of ${CE}_{Loss}$ and $Balance_{Loss}$. ]{\includegraphics[width=1\linewidth]{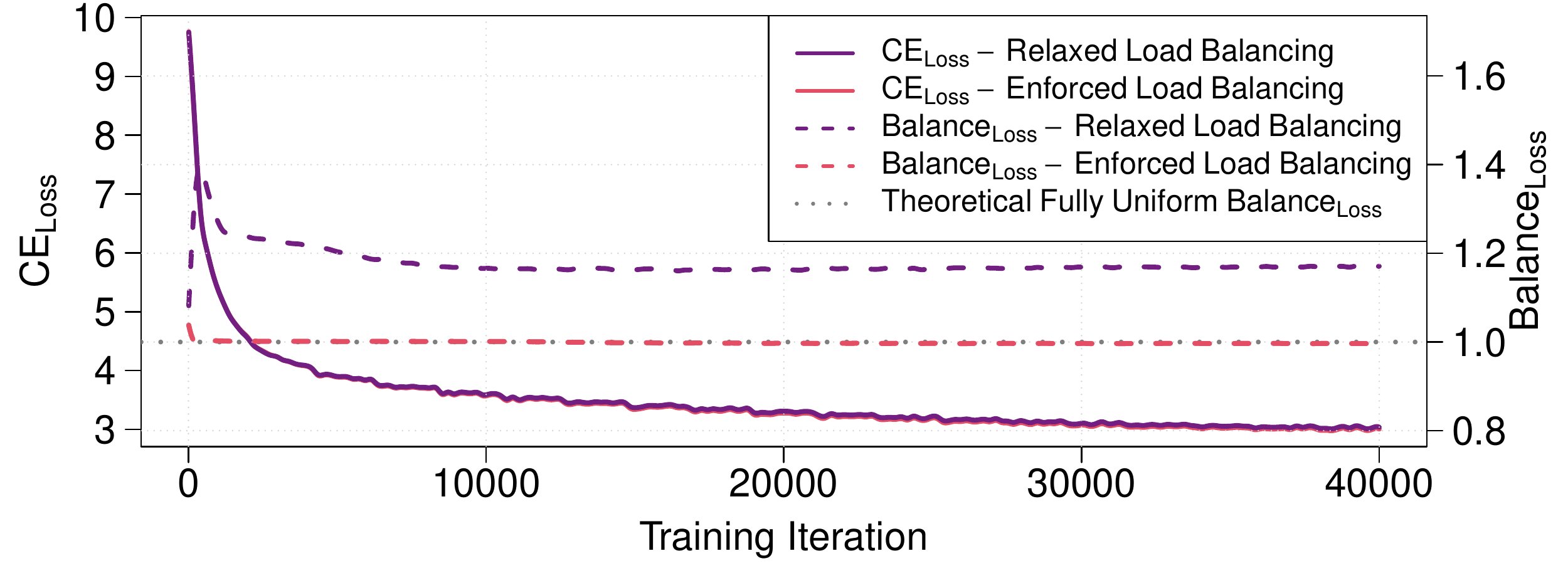}\label{balance_loss}}
\newline
\subfloat[Comparison of MoE Runtime across different batch sizes.]{\includegraphics[width=1\linewidth]{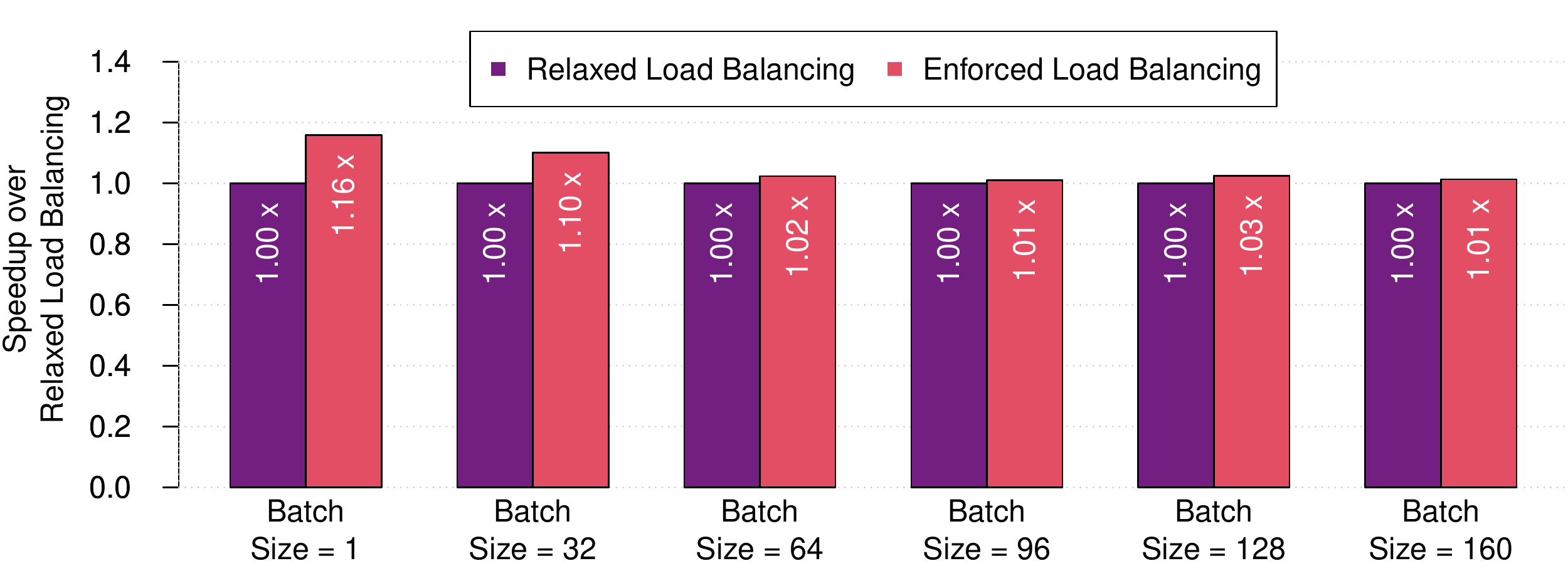}\label{balance_runtime}}
\caption{Impact of relaxing or enforcing the balance loss on training flow as well as MoE runtime.}
\label{balance_exp}
\end{figure}

Figure~\ref{balance_loss} compares the Phase 2 training progress of a Transformer-XL architecture with multiple MoE layers under two scenarios: (1) when the $Balance_{Loss}$ term is excluded from the loss function (\emph{Relaxed Load Balancing}), and (2) when the loss function includes the $Balance_{Loss}$ term  (\emph{Enforced Load Balancing}). From the figure, we notice that trends for the ${CE}_{Loss}$ term are similar in both scenarios, highlighting the fact that overall accuracy of the network is unaffected by load balancing constraints.
%However, enforcing the load balancing across the experts could result in a speedup of up to $1.16\times$ in the runtime of the MoE layers (Figure~\ref{balance_runtime}).
From our experiments, we also notice that a balanced load improves the runtime of MoE layers by reducing tail latency - we illustrate this in Figure~\ref{balance_runtime}. Here, we notice a runtime speedup of up to $1.16\times$ for MoE layers when load balancing is enforced.

%% file: tex/results.tex
\section{Evaluation}
%\salar{We could consider moving section 3.2 or 3.4 to the appendix if we are short of space. I would rather keep the section 3.4, since it provides supporting result for dynamic loss contribution.}

We evaluate \paper on two real-world language modeling tasks and compare the performance of the latency-optimized networks to other state-of-the-art efficient Transformer models.
We also provide a detailed analysis of the impact of using our dynamic loss formulation.

%In this section, we evaluate \paper on two language modeling tasks, we compare the performance of different search blocks, analyze the potential latency savings by applying the 2-step search methodology, and discuss the results of dynamic loss settings with different target latencies.
\subsection{Methodology}
\label{sec:methodology}
We use Transformer-XL (TXL) Base on the WikiText-103 (WT103) and enwik8 datasets as our baseline networks. The backbone architecture for both datasets uses a model dimension of 512 and an interleaved pattern of multi-head attention (MHA) with 8 heads and feed-forward layer (FFLs) with an inner dimension of 2048. The total number of blocks (MHA/FFL) is 24 and 32 for enwik8 and WT103, respectively \footnote{The number of MHA/FFL blocks is $2\times$ of the number of Transformer blocks.}. 
The search space for phase 1 includes: (1) Skip connection, (2) MHA with 1, 2, 4, or 8 heads, (3) FFL with inner dimension of 2048, and (4) MoE FFL with inner dimension of 2048, 8 experts, where each token is processed by either 1 or 2 experts ($Top_K = 1\ or\ 2$). 

%\begin{itemize}
%    \setlength\itemsep{0em}
%    \item Skip Connection
%    \item Multi-head Attentions with 1, 2, 4, or 8 heads
%    \item FFL with inner dimension of 2048
%    \item MoE with inner dimension of 2048, 8 experts, whereas each token is processed either by 1 or 2 experts ($Top_K = 1\ or\ 2$)
%\end{itemize}

To evaluate the performance of \paper, we compare the latency and accuracy of the optimized models with the baseline TXL model and two prior papers: Sandwich Transformer~\cite{press2019improving} and PAR Transformer~\cite{mandava2020pay}. 
The design space is explored using \paper's 2-phase methodology (described in more detail in Section~\ref{sec:search}) with target latencies ranging from $50\%$ to $95\%$. 

All training is performed on a node with 8 NVIDIA V100 GPUs. We use the settings published by NVIDIA for hyper-parameters~\cite{catalog}. The exact hyper-parameters used for each dataset are: 
\begin{itemize}
    \setlength\itemsep{0em}
    \item \textbf{WikiText-103 - Network Weights (Phase 1 and 2):} JITLamb optimizer, learning rate of 0.01, batch size of 256, target and memory length of 192, dropout rate of 0.1 for non-MoE layers and 0.2 for MoE layers, and 40000 iterations.
    \item \textbf{WikiText-103 - Architecture Weights (Phase 1):} Adam optimizer, learning rate of 0.01, initial temperature of 5 for the Gumbel Softmax, and temperature annealing rate of 0.6.
    \item \textbf{enwik8 - Network Weights (Phase 1 and 2):} JITLamb optimizer, learning rate of 0.004, batch size of 64, target and memory length of 512, dropout rate of 0.1 for non-MoE layers and 0.3 for MoE layers, and 120000 iterations.
    \item \textbf{enwik8 - Architecture Weights (Phase 1):} Adam optimizer, learning rate of 0.01, initial temperature of 5 for the Gumbel Softmax, and temperature annealing rate of 0.7.
\end{itemize}
%We also use lookup tables filled with the latency of individual blocks, profiled using a batch size and $target\_len$ of 64 on NVIDIA A100 GPU.

\subsection{Accuracy and Performance Trade-offs}
\label{Results_accuracy_perf_section}

\begin{table}[tb]
\centering
\scalebox{0.9}{
\renewcommand{\arraystretch}{1}
\begin{tabulary}{1cm}{*5c}
\hline
\multicolumn{1}{c}{\textbf{Model}} & \multicolumn{2}{c}{\textbf{wt103 (PPL)}} & \multicolumn{2}{c}{\textbf{enwik8 (BPC)}} \\ \hline
{} & Dev & Test & Dev & Test \\
Transformer-XL Base & 22.7 & 23.4 & 1.114 & 1.088 \\ 
Sandwich Transformer-XL & $22.6^*$ & - & 1.107 & 1.083  \\ 
PAR Transformer-XL & $22.7^*$ & - & 1.121 & 1.119 \\ 
\paper Transformer-XL & 22.5 & 23.5 & 1.109 & 1.083 \\ \hline
\end{tabulary}
}
\caption{Accuracy comparison of \paper with prior work and baselines (scores marked with $*$ are referenced). Lower is better for both PPL and BPC metrics.}
\label{acc_results}
\end{table}

Table~\ref{acc_results} lists the accuracy numbers obtained by \paper and compares them with the baseline architectures. We notice that all the TXL variants, including ones produced by \paper, maintain baseline accuracy levels. We provide a detailed comparison of the different architectures in Appendix~\ref{appendix_evaluated_architectures}.

\begin{figure}[tb]
\subfloat[WT103]{\includegraphics[width=1\linewidth]{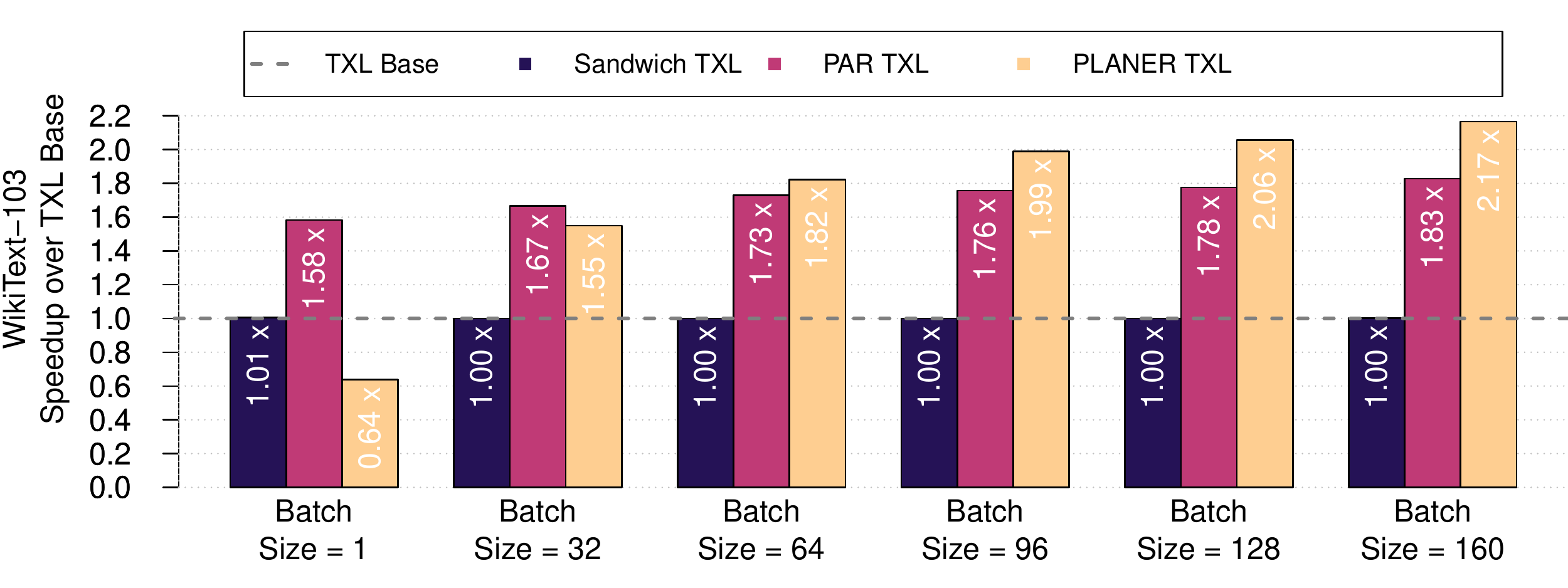}\label{wt103_latency}}
\newline
\subfloat[enwik8]{\includegraphics[width=1\linewidth]{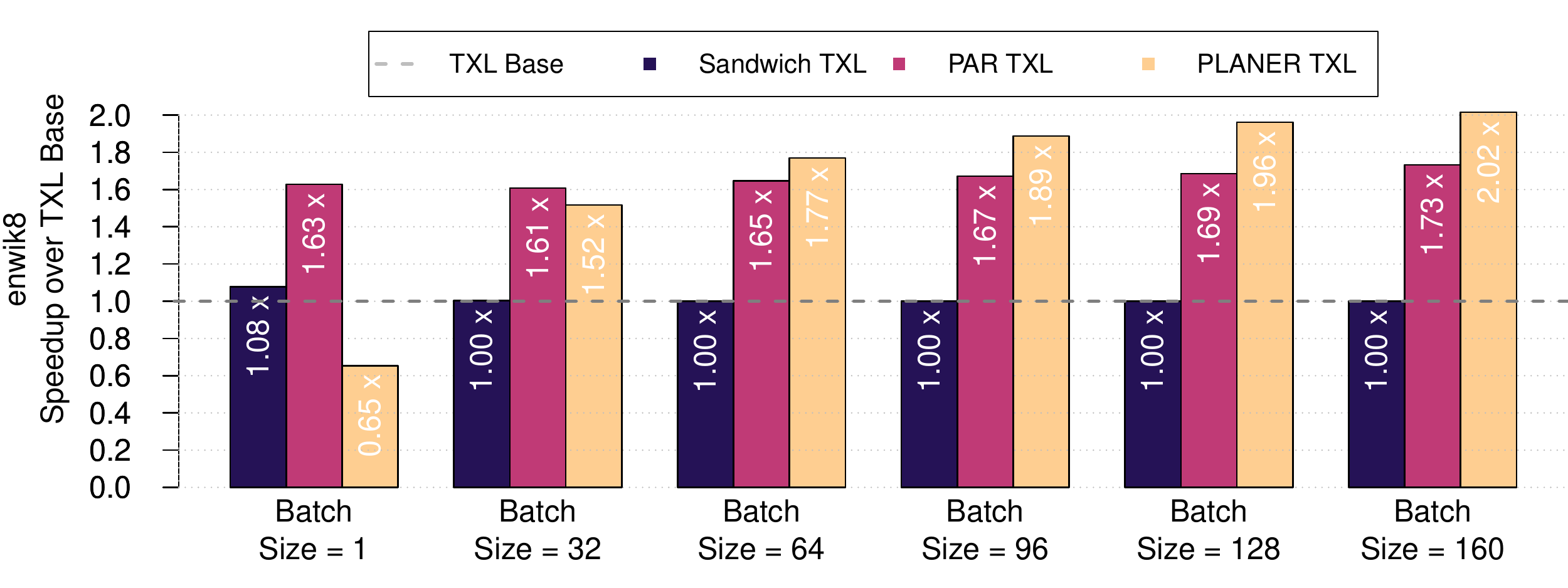}\label{enwik8_latency}}
\caption{Speedups obtained by \paper w.r.t.~various baselines across different batch sizes, profiled on NVIDIA A100.}
\label{both_latency}
\end{figure}

%As for the speedups, we evaluate each individual architecture across various batch sizes.
Figure~\ref{both_latency} shows the speedups obtained by \paper and the various baselines (described in Section~\ref{sec:methodology}) across both datasets and varying batch sizes. From the Figure, we notice that \paper provides speedups of over $2\times$ at larger batch sizes.
On smaller batch sizes, PAR Transformer outperforms \paper; this is primarily due to the unoptimized MoE layers used in our current implementation. Specifically, our current implementation computes the outputs of each expert sequentially, where a batch of sequences with $N$ tokens are sequentially processed in mini-batches of size
$\frac{Top_K \times N}{Experts}$. This consequently leads to under-utilization of the compute units. 

\begin{figure}[t]
  \centering
  \includegraphics[width=1\linewidth]{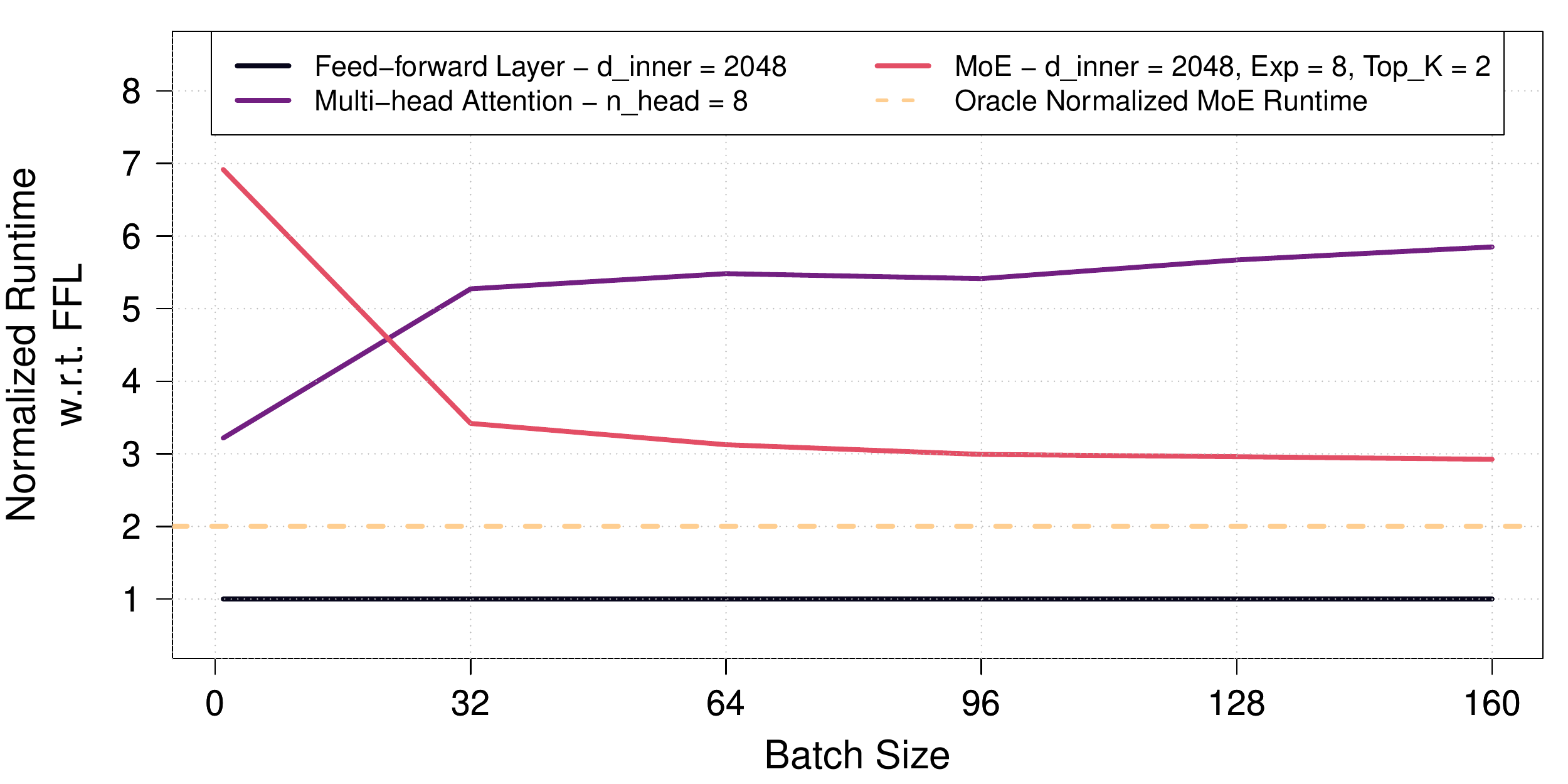}
  \caption{Runtime comparison of FFL, MHA, and MoE layers across different batch sizes normalized to FFL runtime.}
  \label{ffl_moe_attn_comp}
\end{figure}

Figure~\ref{ffl_moe_attn_comp} provides a more detailed overview of the current deficiencies in the sequential implementation of the MoE layers; here, we provide a runtime comparison of the FFL, MHA, and MoE layers across different batch sizes normalized with respect to FFL runtime. At lower batch sizes, MoE layers have an overhead of $7\times$ over FFL, which is also higher than the MHA layers. However, as batch size increases, GPU resource utilization goes up, consequently decreasing the overhead of MoE layers to less than $3\times$.
The oracle implementation (dashed orange line in Figure) shows the theoretically optimal runtime of the MoE layer. Since we use a $Top_k$ value of $2$ (viz., each input token is processed by 2 experts), we notice a corresponding $2\times$ runtime overhead over the baseline FFL. Note that the oracle runtime does not take overheads related to gate function evaluation and the gathering/scattering of tokens across experts into account - the real-world runtime is thus likely to be higher.
We are currently working on a more optimized parallel implementation of MoE layers, which will help plug this performance gap across various batch sizes.
%an see that in large batch size settings, the aggressive pruning of \paper results in more than $2\times$ speedup gains. However, when the batch size is not high enough, PAR Transformer outperforms the architecture explored by \paper. This phenomenon could be explained by the non-optimal implementation of MoE computation that is currently deployed. In the current setup, the computation of each expert is done sequentially. This means that to compute a 
%batch of sequences with an overall $N$ tokens, we sequentially process them in mini-batches with size of $\frac{Top_K \times N}{Experts}$. However this sequential process would lead to inefficiency in the smaller batch sizes due to the under-utilization of the compute units, compared to processing all tokens in parallel using a single large batch size. The implementation of an efficient and parallel MoE computation is currently under progress, which could help to further improve the gains across the batch size spectrum.  
 %The missing data points in the table correspond either to the lack of architecture referenced in the prior works, or failure to generate an iso-accuracy architecture.

\subsection{Comparison to Iso-parametric Setting}
We also compare \paper to an iso-parameter setup, which replaces the MoE with a scaled FFL in the search space. The scaled FFL has an inner dimension of 16384, which results in the same number of parameters as the MoE with 8 experts. The goal of the iso-parameter experiment is to analyze the effectiveness of different model scaling solutions in compensating for accuracy drops caused by aggressive attention pruning.

\begin{figure}[t]
  \centering
  \includegraphics[width=1\linewidth]{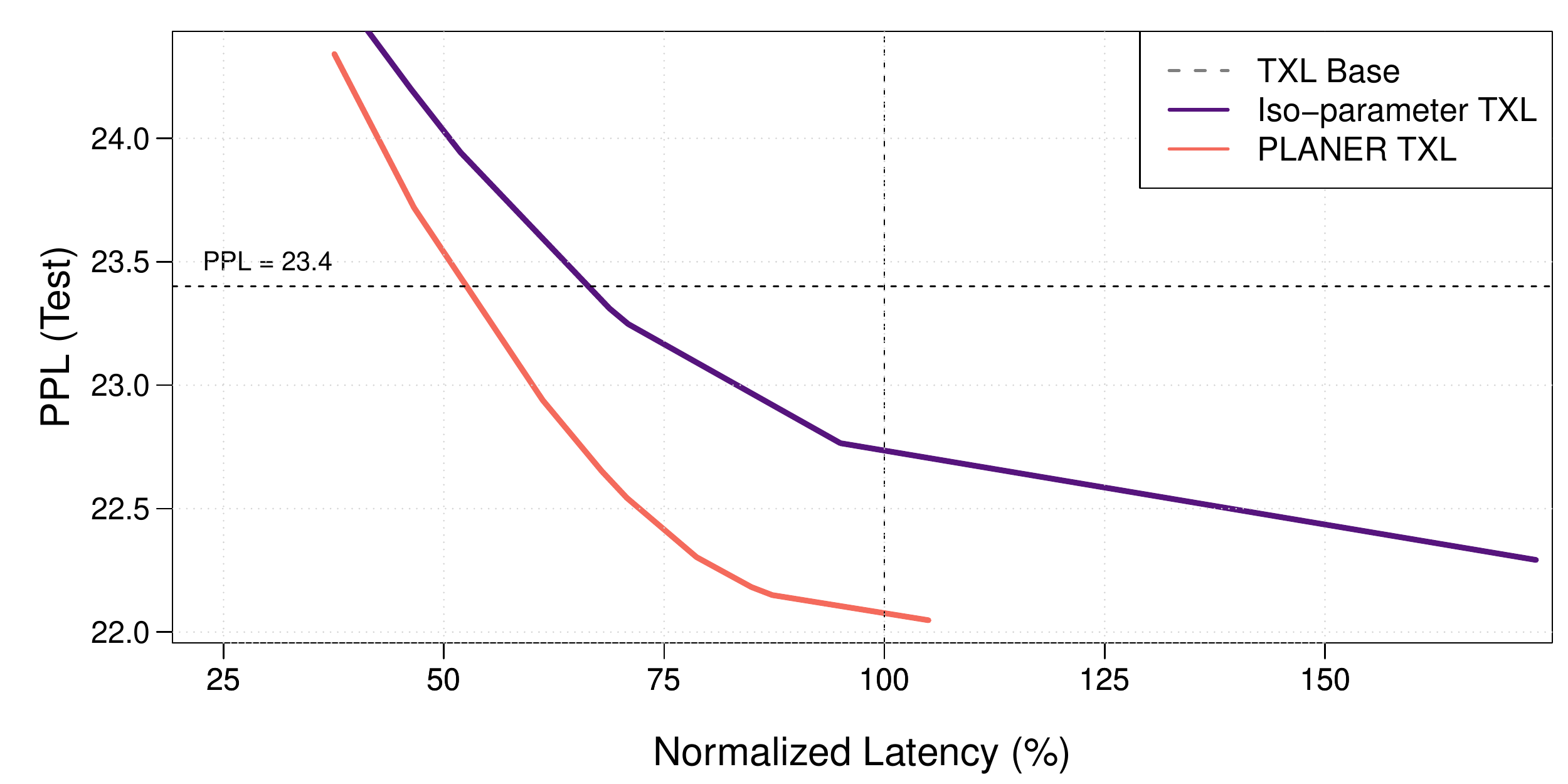}
  \caption{Comparison of the Pareto frontiers of the optimized architectures obtained by \paper for MoE and Iso-parameter scaled FFL setups.}
  \label{wt103_pareto}
\end{figure}

Figure~\ref{wt103_pareto} presents the comparison of the Pareto frontiers of the architectures obtained by \paper with different latency targets on the WikiText-103 dataset. From the Figure, we clearly notice that the use of MoE layers results in higher performance architectures across the board at different accuracy levels. Further performance benchmarking reveals that scaled FFL layers are at least $2\times$ slower than our (relatively unoptimized) MoE layers and actually approach the runtime of the much slower MHA layers with 8 heads. Naively scaling up the size of FFLs is thus not an ideal option for either improving accuracy or performance.

\subsection{Validating Estimated and End-to-end Runtime}

\begin{figure}[t]
\subfloat[Target vs estimated]{\includegraphics[width=0.5\linewidth]{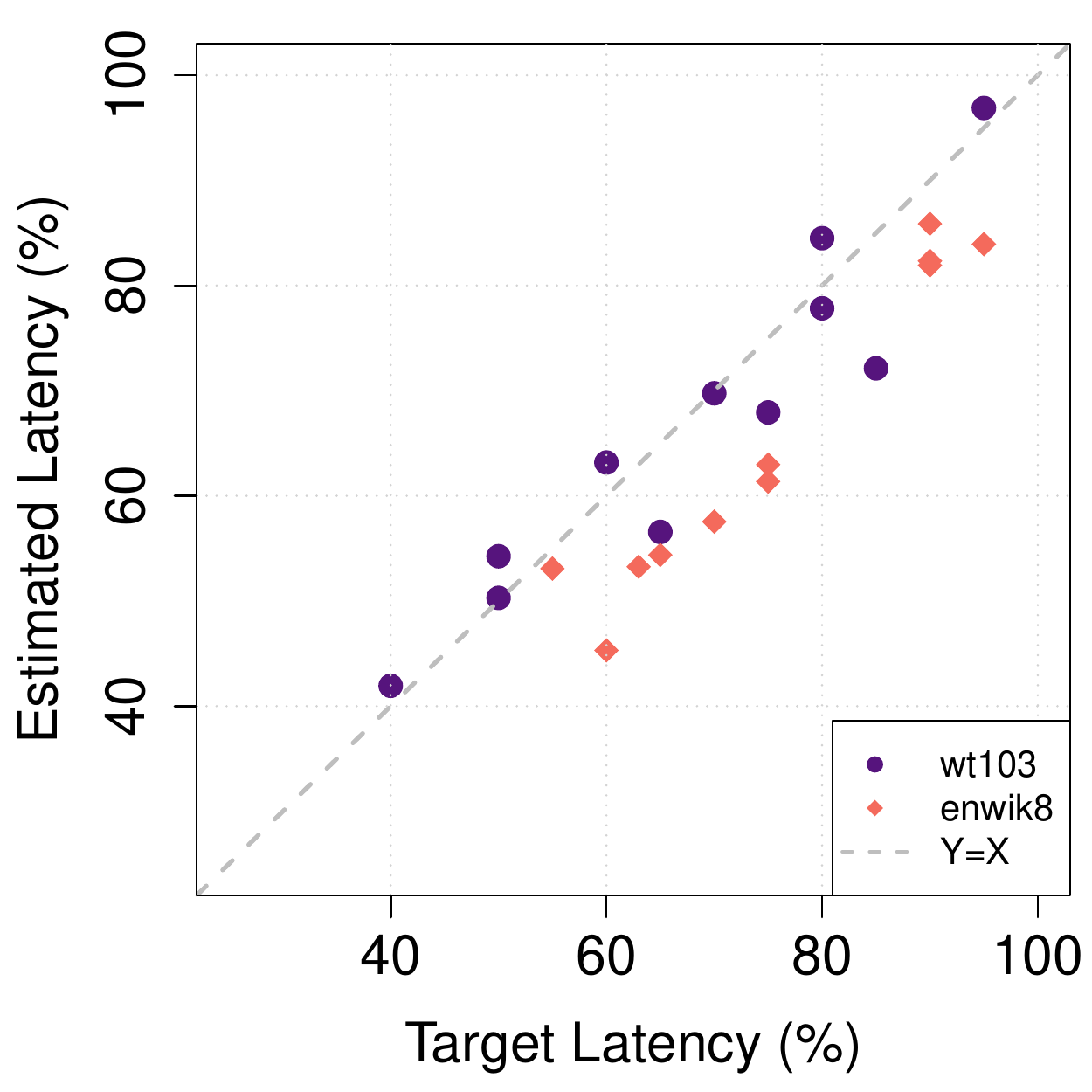}\label{target_project}}
\subfloat[Estimated vs end-to-end]{\includegraphics[width=0.5\linewidth]{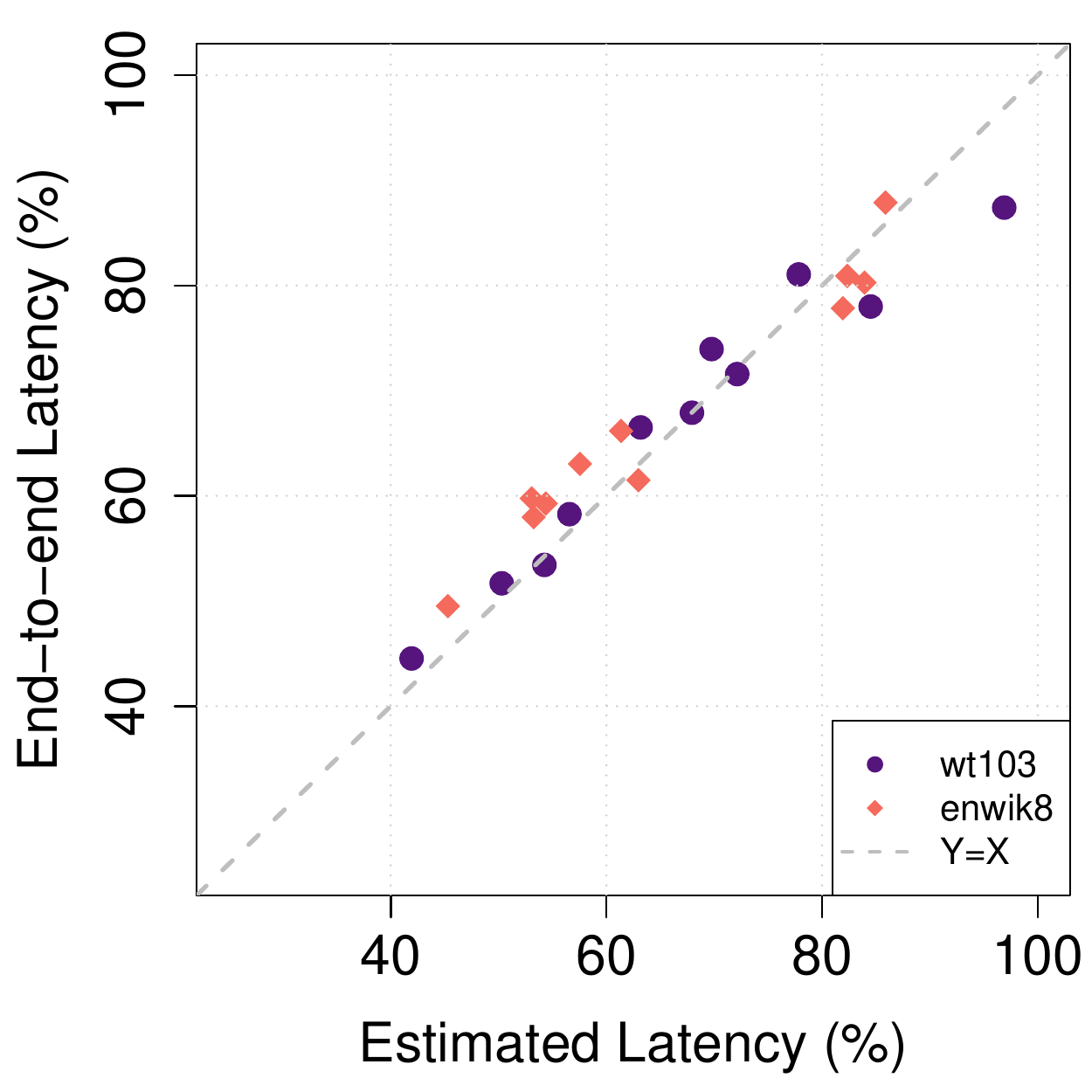}\label{project_end}}
\caption{Correlation between target, estimated, and end-to-end latency.}
\end{figure}

In this section, we analyze the performance of the dynamic latency loss used in Phase 1.
Figure~\ref{target_project} shows the correlation between input target latency and the estimated latency of the architectures sampled at the end of Phase 1, while
Figure~\ref{project_end} shows the correlation between estimated latency and profiled end-to-end latency.
We make two important observations from the figures: (1) our dynamic latency loss formulation successfully steers the NAS towards architectures that match the input target latency, and (2) the latency estimated in Equation (2) is highly correlated with real-world latency, making it an appropriate option for \paper's Phase 1 search. 

\subsection{Repeatability Evaluation}
\begin{figure}[t]
\subfloat[WT103]{\includegraphics[width=0.5\linewidth]{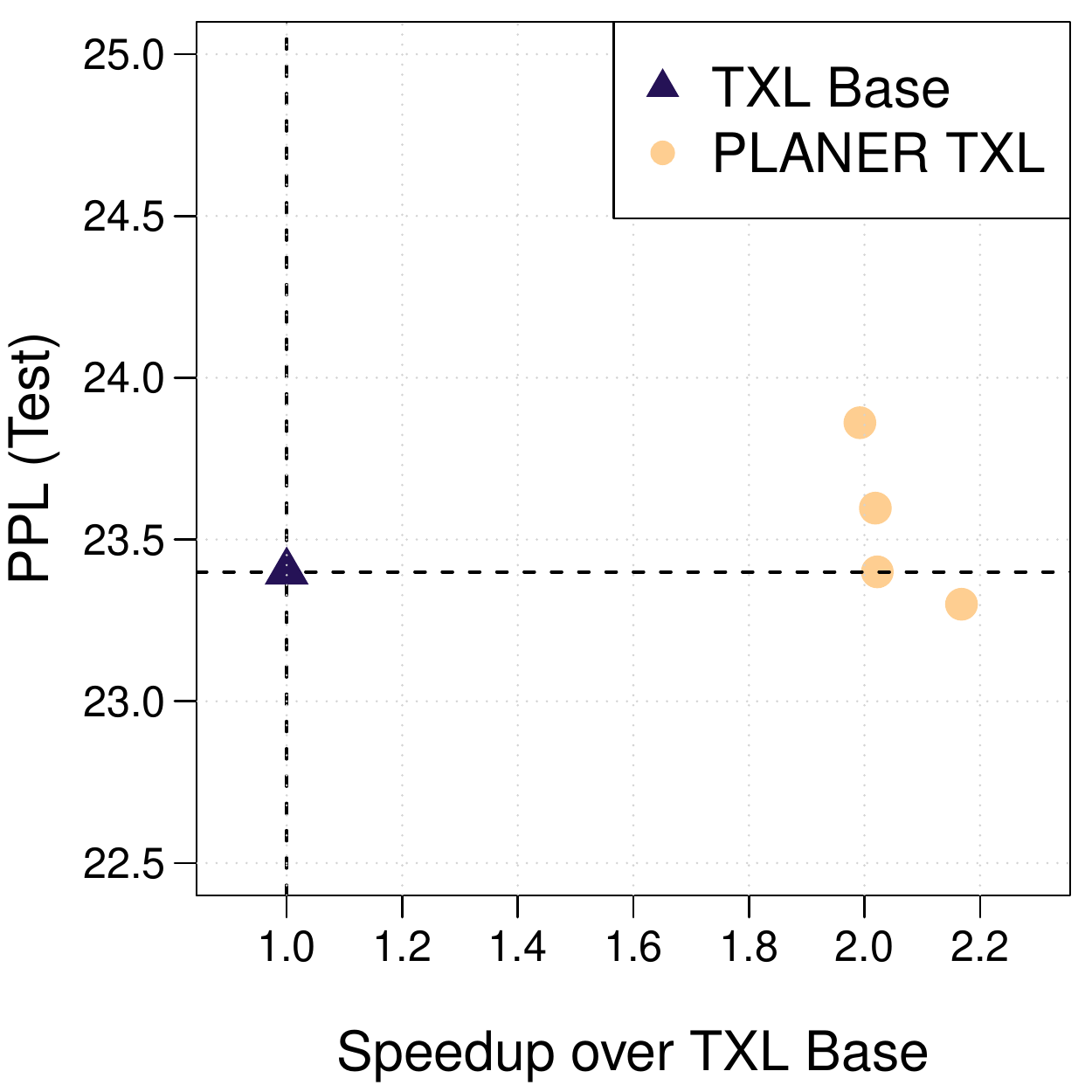}\label{repeatability_wt103_latencies}}
\subfloat[enwik8]{\includegraphics[width=0.5\linewidth]{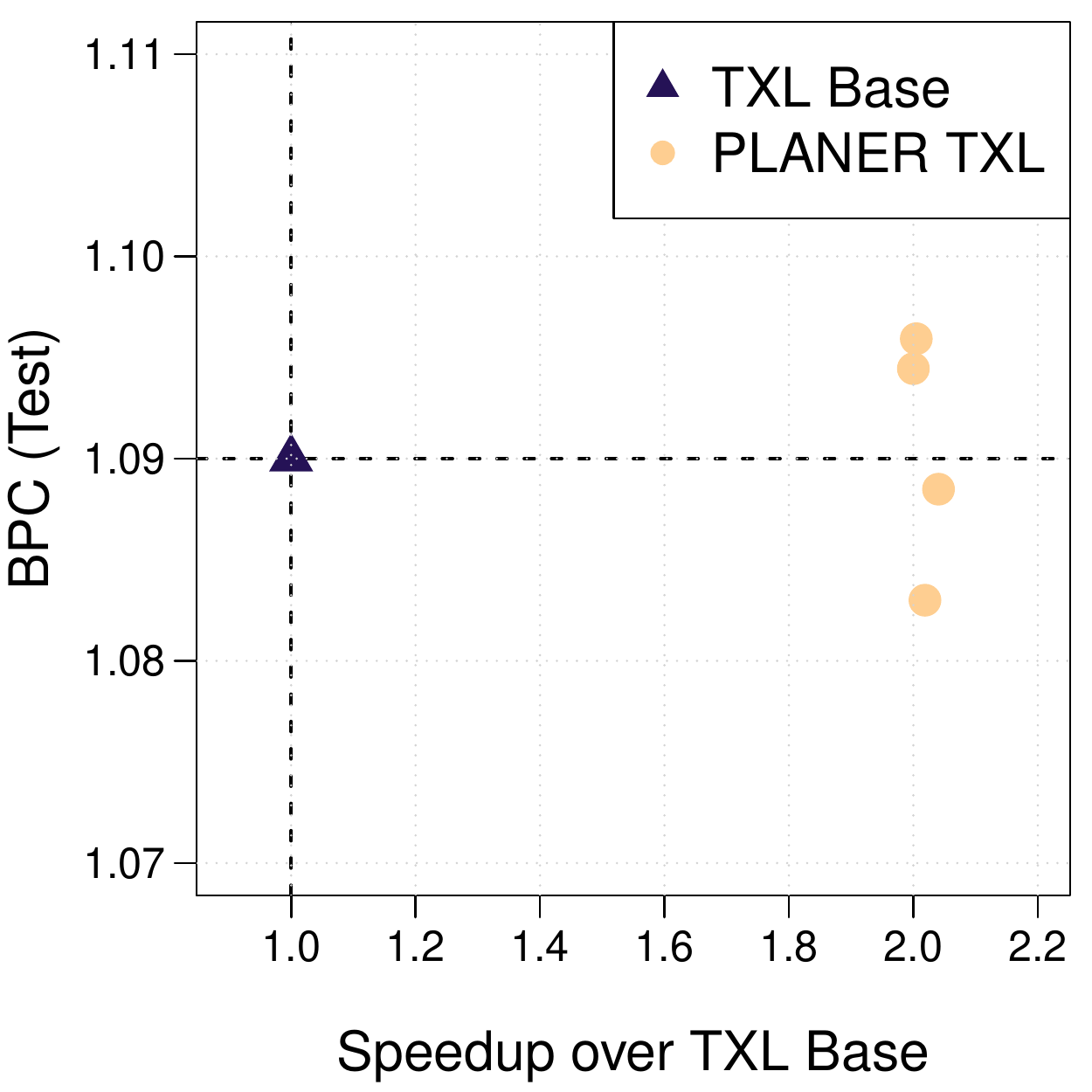}\label{repeatability_enwik8_latencies}}
\caption{Speedup and accuracy results for the repeatability experiment.}
\label{both_repeatability}
\end{figure}

To evaluate and validate the reproducibility of our experiments, and observe any potential variations in the final architectures, we also repeat the \paper optimization of the architectures evaluated in Section~\ref{Results_accuracy_perf_section}. For this experiment, we keep all hyper-parameters fixed, but repeat \paper's search process four times.
Figure~\ref{both_repeatability} presents the achieved accuracy and speedup numbers from our experiment. We notice that all the accuracy values are within $0.5\%$ of the baseline, with speedups consistently over $2\times$. The variations in the final architectures across the two datasets are presented in Appendix~\ref{appendix_repeatability_architecture}. Although the architectures do not match exactly, we notice a strong similarity in the number of heads in the attention layers. We also noticed that MoE layers tend to be concentrated towards the end of the networks across both datasets.

%We compare the target latencies with the estimated latencies of the architectures sampled at the end of phase-1. We also measure the end-to-end latency of the explored architectures to evaluate the correlation of it with respect to the estimated latencies.

%* Why do MoE layers preserve accuracy?
%  * Input space partitions
%  * More effective parameters vs. constant per-sample parameters
%* Discussion on the impact of load balancing.
%* Discussion on the inference approach of MoE layers, i.e., limitations of sequential vs opportunities in parallel. 

%% file: tex/related.tex
\section{Related Work}
The introduction of the Transformer family of networks has overhauled the domain of NLP. These attention-based architectures have been shown to outperform their LSTM-based counterparts both in terms of effectively capturing time dependencies~\cite{vaswani2017attention} as well as inference latency~\cite{shi2021emformer}. The general architecture of these models consists of multiple Transformer blocks, where each Transformer block consists of multi-head attention(s) and feed-forward layers.

Recent work has introduced Mixture-of-Expert (MoE) layers within networks to decompose tasks into sub-tasks, where experts could be trained on individual sub-tasks~\cite{masoudnia2014mixture}. One motivation behind this idea is to dynamically partition the input space, with experts getting specialized on individual partitions.  
Recent work has also studied the application of MoE layers to efficiently increase the model capacity of Transformer-based architectures~\cite{shazeer2017outrageously, lepikhin2020gshard, fedus2021switch, he2021fastmoe}. These sparsely-activated architectures have been shown to achieve accuracy gains without the proportional increase in computation compared to traditional scaling of network parameters~\cite{raffel2019exploring}. While MoE layers have been applied for accuracy improvement and training speed-ups, their use in designing latency-aware architectures have not been explored as thoroughly. 

A separate body of work has also focused on optimizing the performance of Transformer models. In particular, \cite{press2019improving} show that it is possible to achieve better accuracy by redistributing multi-head attention and FFL layers across the network while maintaining the original runtime.
%compared to the vanilla interleaved configuration
PAR~\cite{mandava2020pay} deploys NAS to explore the number and distribution of attention layers (while keeping the same head count) for improved latency. \cite{wang2020hat} prune attention heads and reduce the width of FFLs (while keeping the same backbone as the baseline) using an evolutionary NAS algorithm to design hardware-aware Transformers. While recent work has explored the distribution or configuration of individual (non-MoE) layers in isolation, none of them consider both aspects simultaneously as part of a larger NAS search space. Additionally, as we demonstrate in this paper, the inclusion of MoE layers in the design space can help reduce inference latency further by removing/pruning attention layers more aggressively while maintaining baseline accuracy.

%* Dynamic transformers or MoE + Transformer work

%% file: tex/conclusion.tex
\section{Conclusion}

This paper has presented \paper, an automated system for optimizing the inference latency of Transformer-based networks. \paper employs a two-phase NAS methodology to systematically introduce sparsely activated layers into the given network, and uses a dynamic loss formulation to achieve user-provided latency targets while preserving accuracy. On two real-world NLP models, \paper achieves inference latency reductions of over $2\times$ at iso-accuracy.

%% file: tex/appendix.tex
\section{Evaluated Architectures}
\label{appendix_evaluated_architectures}
\begin{figure}[H]
  \centering
  \includegraphics[width=1\linewidth]{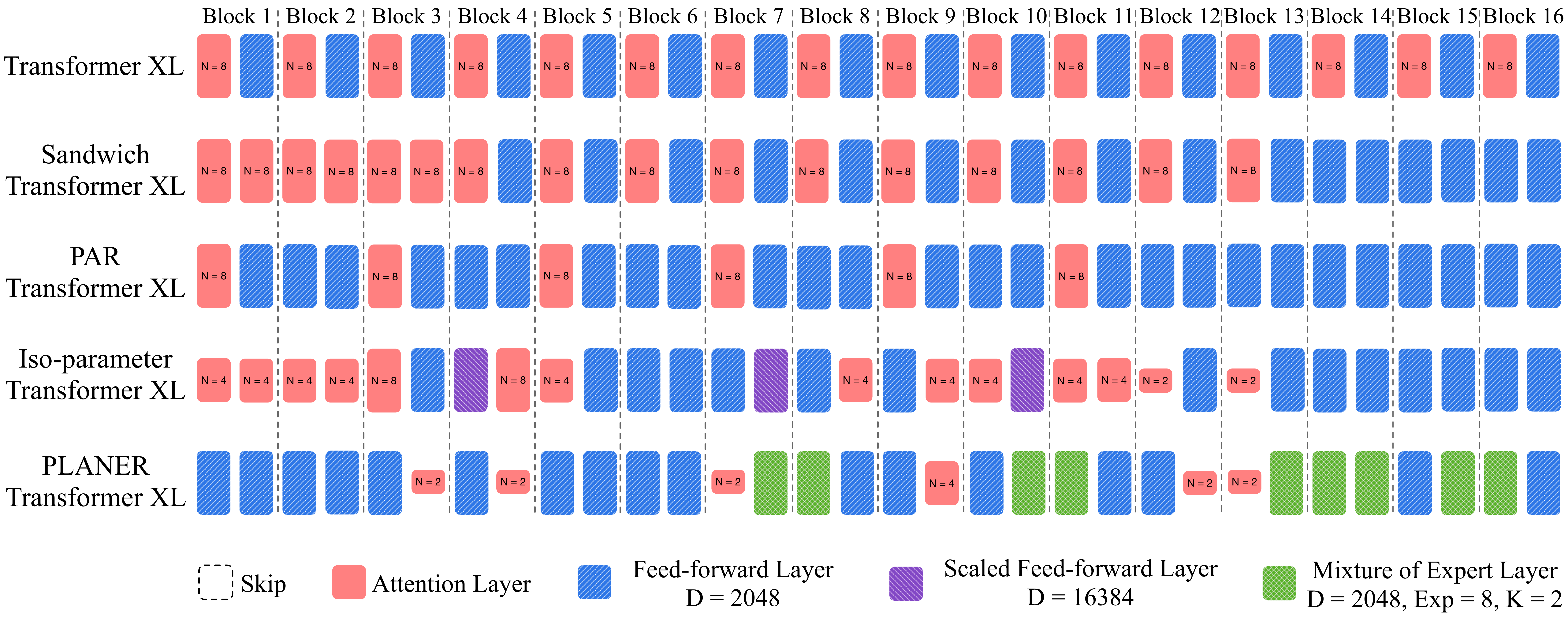}
  \caption{Evaluated architectures for WT103 dataset.}
  \label{all_architectures_wt103}
\end{figure}

\begin{figure}[H]
  \centering
  \includegraphics[width=1\linewidth]{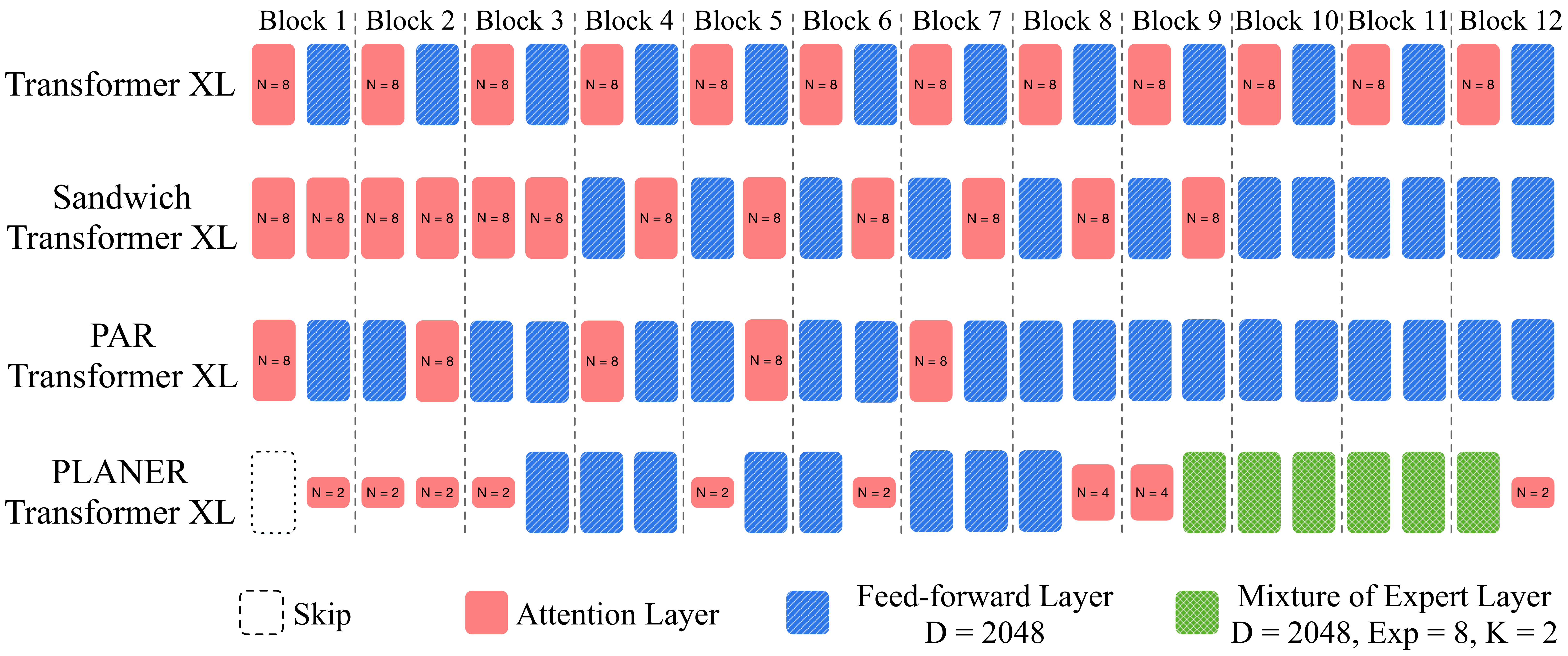}
  \caption{Evaluated architectures for enwik8 dataset.}
  \label{all_architectures_enwik8}
\end{figure}

Figures~\ref{all_architectures_wt103} and~\ref{all_architectures_enwik8} present the detailed architecture of all evaluated models in Section~\ref{Results_accuracy_perf_section}. We notice that PLANER aggressively prunes/skips attention layers, while intelligently introducing sparsely activated layers for accuracy recovery.

\newpage

\section{Repeatability Experiments: Architecture Comparison}
\label{appendix_repeatability_architecture}

\begin{figure}[H]
  \centering
  \includegraphics[width=1\linewidth]{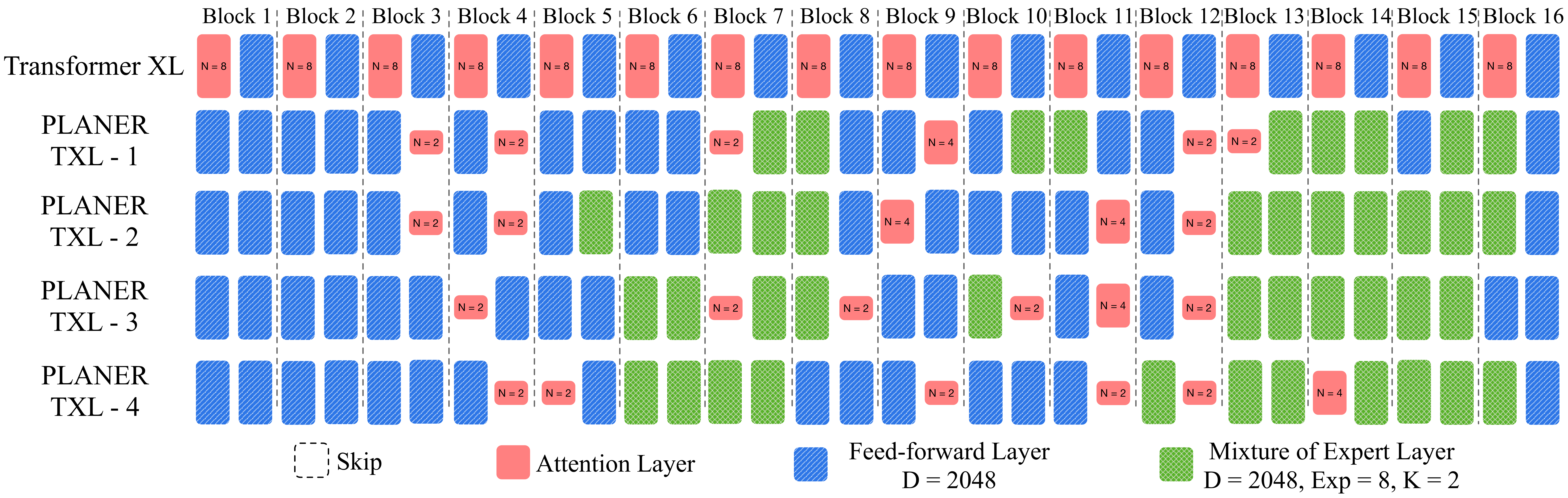}
  \caption{Explored architectures through repeatability experiment on WikiText-103 dataset.}
  \label{repeatability_wt103_architectures}
\end{figure}

\begin{figure}[H]
  \centering
  \includegraphics[width=1\linewidth]{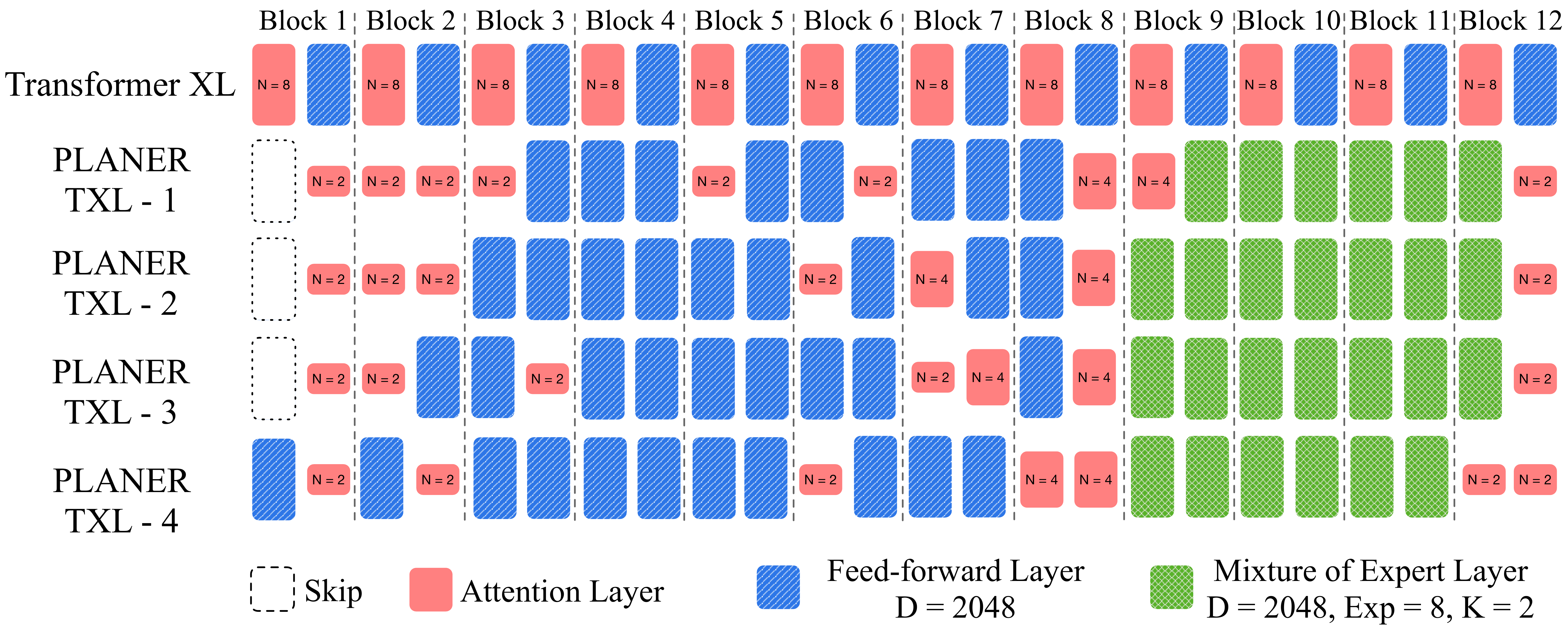}
  \caption{Explored architectures through repeatability experiment on enwik8 dataset.}
  \label{repeatability_enwik8_architectures}
\end{figure}